\documentclass[10pt,twocolumn,letterpaper]{article}

\usepackage{iccv}
\usepackage{times}
\usepackage{epsfig}
\usepackage{graphicx}
\usepackage{amsmath}
\usepackage{amssymb}
\usepackage{bbding} 
\usepackage{subfigure}
\usepackage{bm}
\usepackage{enumerate}
\usepackage{multirow}
\usepackage{multicol}
\usepackage{colortbl}
\usepackage{color}
\usepackage{mathtools}
\usepackage{booktabs}
\usepackage{cuted}
\usepackage{capt-of}

\usepackage[pagebackref=true,breaklinks=true,letterpaper=true,colorlinks,bookmarks=false]{hyperref}

\iccvfinalcopy 

\def\iccvPaperID{8631} 

\ificcvfinal\fi

\begin{document}
	
	\title{ASFlow: Unsupervised Optical Flow Learning with Adaptive Pyramid Sampling}
	
	\author{%
	Kunming Luo$^1$ \quad
	Ao Luo$^1$ \quad
	Chuan Wang$^1$  \quad
	Haoqiang Fan$^1$\quad
	Shuaicheng Liu$^{2,1}$\thanks{Corresponding author} \\\\
	$^1$Megvii Technology \quad
	$^2$University of Electronic Science and Technology of China
	}
	
	\maketitle
	\ificcvfinal\thispagestyle{empty}\fi
	
	\begin{abstract}
		We present an unsupervised optical flow estimation method by proposing an adaptive pyramid sampling in the deep pyramid network. Specifically, in the pyramid downsampling, we propose an Content Aware Pooling (CAP) module, which promotes local feature gathering by avoiding cross region pooling, so that the learned features become more representative. In the pyramid upsampling, we propose an Adaptive Flow Upsampling (AFU) module, where cross edge interpolation can be avoided, producing sharp motion boundaries. Equipped with these two modules, our method achieves the best performance for unsupervised optical flow estimation on multiple leading benchmarks, including MPI-SIntel, KITTI 2012 and KITTI 2015. Particuarlly, we achieve EPE=1.5 on KITTI 2012 and F1=9.67\% KITTI 2015, which outperform the previous state-of-the-art methods by 16.7\% and 13.1\%, respectively.
	\end{abstract}
	
	\section{Introduction}
	Optical flow estimation is a long lasting research topic since proposed by Horn and Schunck~\cite{horn1981determining}. It is a fundamental technique for many computer vision applications~\cite{Jiang_2018_CVPR,largemotion_network_design_iccv2017,Simonyan2014}. Early methods optimize the pre-defined energy functions with various assumptions and constraints~\cite{Sun2010,Sun2010nips,Sun2014cvpr,Sun2016}. The learning-based optical flow methods become more popular than the traditional variational-based counterparts due to their leading performances in benchmark evaluations and real-time inference speed.
	
	The DNN-based methods can be classified into supervised~\cite{Flownet_flyingchairs,spynet2017,LiteFlowNet,pwc_net,irrpwc} and unsupervised~\cite{unflow_2018aaai,Pengpeng2019,Epipolar_flow_2019cvpr,liu2020learning,jonschkowski2020matters} approaches. The training of supervised methods require the ground-truth flow labels, which is hard to obtain. As a result, these models are primarily trained on large-scale synthetic datasets~\cite{Flownet_flyingchairs,Butler2012}, because obtaining ground-truth annotations for real-world scenarios is prohibitively expensive. Consequently, the supervised methods may suffer from domain transfer problems, where the synthesized images are different from the real ones.  
	
	In unsupervised methods, the ground-truth annotations are not necessary. The photometric loss is optimized by warping one image to the other with predicted optical flows. Without the label guidance, occlusions and motion boundaries need special attentions during the unsupervised training process~\cite{jonschkowski2020matters,luo2020occinpflow}. 
	
	\begin{figure}[t]
		\centering
		\subfigure[Feature similarity map with and w/o our content aware pooling (CAP). ]{
			\includegraphics[width=0.98\linewidth]{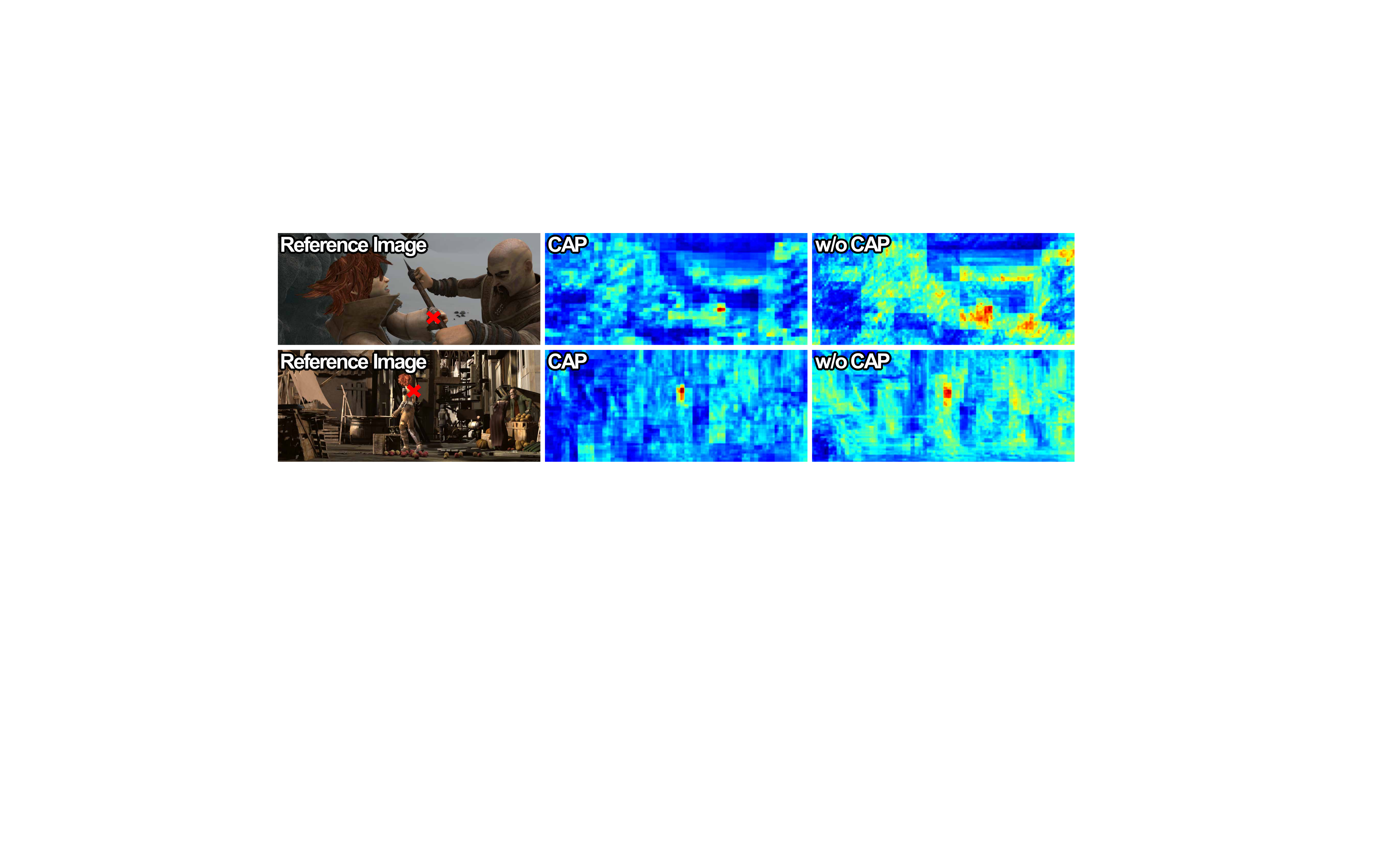}
			\label{fig:teaser_a_feature_vis}
		}
		\centering
		\subfigure[An example from Sintel Clean benchmark.]{
			\includegraphics[width=0.98\linewidth]{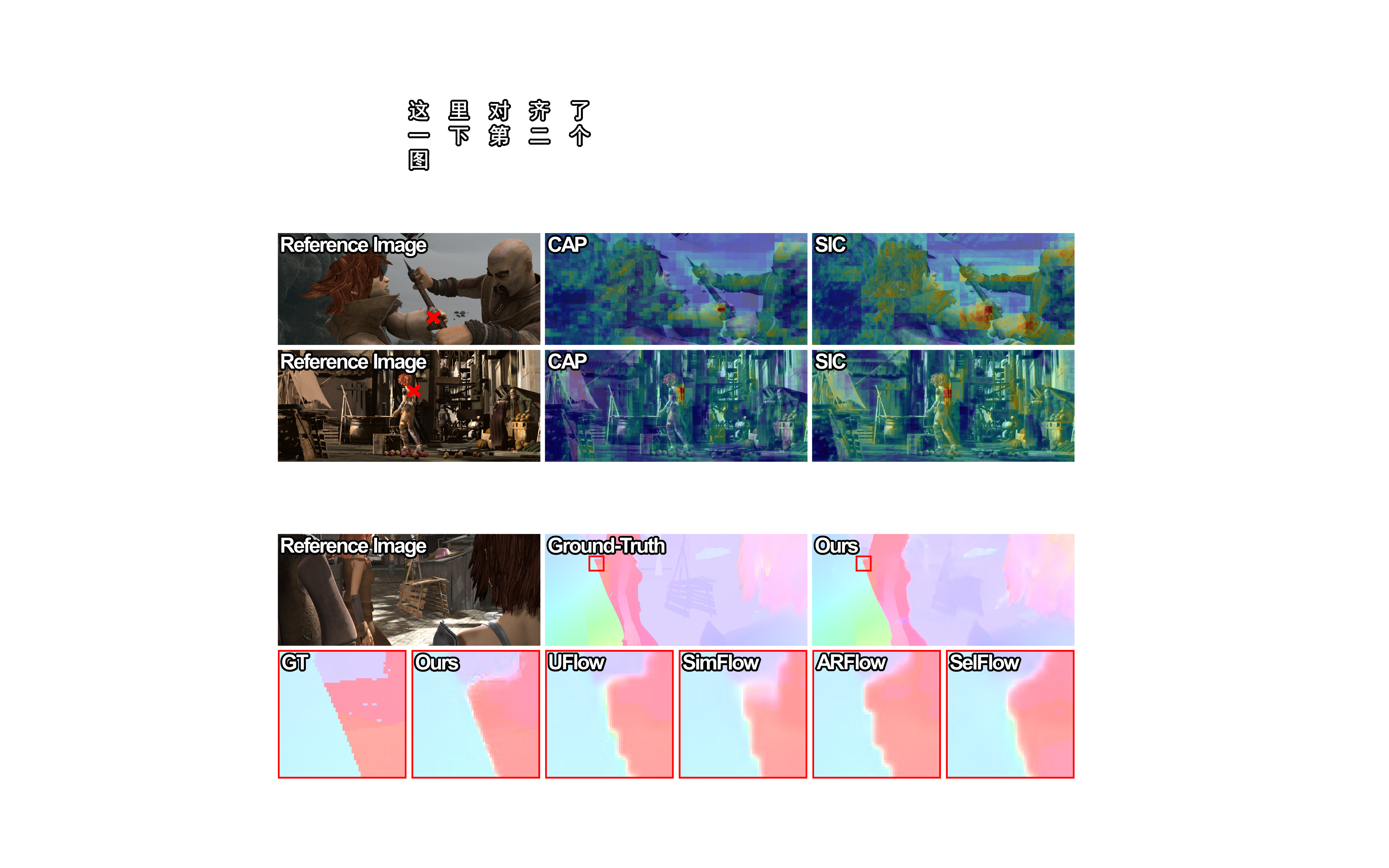}
			\label{fig:teaser_b_upsample_vis}
		}
		\caption{Some examples from Sintel Clean benchmark. (a) With our proposed CAP, the learned features are more representative. (b) Compared with previous unsupervised methods, UFlow~\cite{jonschkowski2020matters}, SimFlow~\cite{simFlow2020eccv}, ARFlow~\cite{liu2020learning}, and SelFlow~\cite{Liu2019CVPR}, our approach produces sharper and more accurate results at motion boundaries.
		}\label{fig:teaser_sintel_clean}
	\end{figure}

	\begin{figure*}
		\centering
		\includegraphics[width=0.98\linewidth]{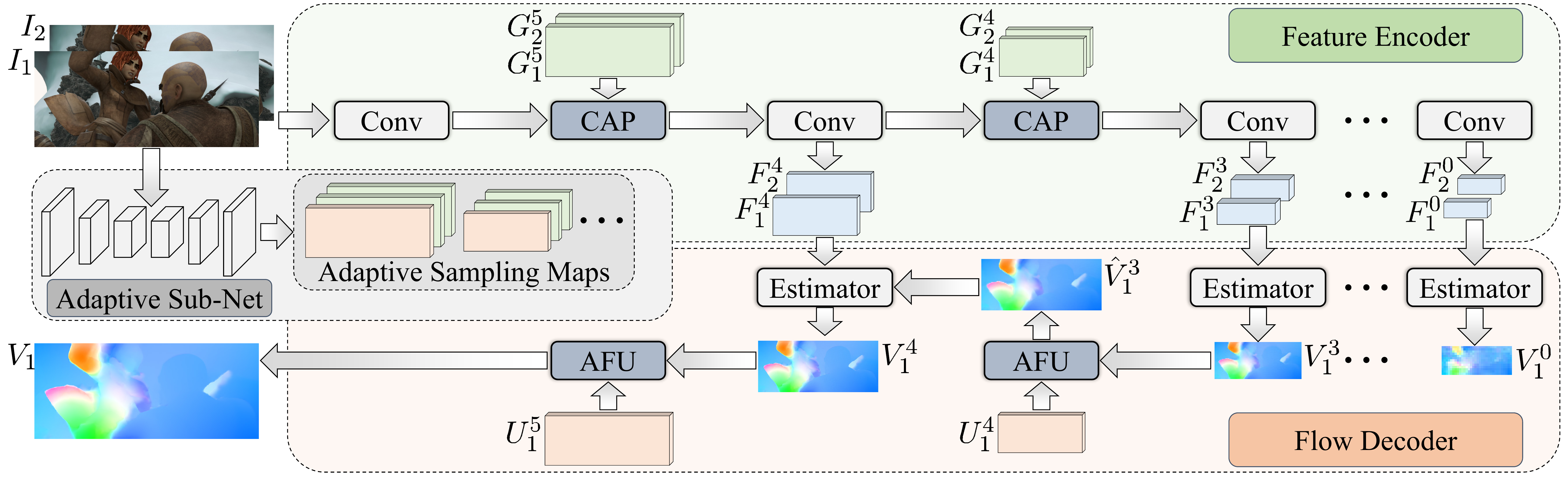}
		\caption{Illustration of our network, where `Conv' represents a convolutional block that contains two convolution layers with kernel size $3$ and stride $1$, `Estimator' denotes the conventional optical flow estimator, `CAP' is the proposed Content Aware Pooling module, and `AFU' is the proposed Adaptive Flow Upsampling module. 
		}\label{fig_algo_our_pipeline}
	\end{figure*}
	
	The pyramid structure is popular in the optical flow learning, where global and local motions can be estimated in a coarse-to-fine manner. We notice that there are two components that should be improved in the pyramid structure~\cite{pwc_net,irrpwc}. One is related to the pyramid downsampling and the other is the upsampling.   
	
	In the process of pyramid downsampling, the network adopts striding in convolution (SIC) or the pooling to decrease the feature sizes. However, the striding or pooling is fixed with a rectangular size, which may not be optimal for the feature information gathering. Considering that, a rectangle may span different image regions, where multiple irrelevant values are forced to gather together, picking one of them may not be optimal, yielding values that are less representative. On the other side, in the pyramid upsampling, the flows are interpolated from coarse-to-fine. However, such an interpolation may cross image edges, resulting in the blur effects in the estimated flows. Even worse, such errors will be propagated and aggregated when the scale becomes finer.
	
	Based on the above observations, we propose an Adaptive Pyramid Sampling approach to upgrade the pyramid network structure, including a \emph{Content Aware Pooling} (\textbf{CAP}) module for the pyramid downsampling and an \emph{Adaptive Flow Upsampling} (\textbf{AFU}) module for the pyramid upsampling. The CAP can automatically group image features, such that the similar features can be gathered locally before the downsampling. With our CAP, the learned features become more representative, so as to promote the overall performance. On the other side, the AFU module interpolates the flows adaptively, where cross edge interpolation can be avoided, leading to sharper flows at motion boundaries. Specifically, in the AFU, we propose a \emph{sampling regularization loss} to constrain the learned adaptive sampling maps, where the upsampled flow fields can better fit the object boundaries. 
	
	Fig.~\ref{fig:teaser_sintel_clean} provides some visualization results on Sintel Clean dataset.  Specifically, Fig.~\ref{fig:teaser_a_feature_vis} shows some feature similarity maps. We extract features from source and target images. We choose one feature vector at a position (marked in red cross) from the source image and calculate its similarity with all features at the target image. We plot the similarity as a heat map, where high similarity values are depicted in red. As seen, with our CAP, the feature at the `red cross' is quite different from the features at the other places. In contrast, without our CAP, features at many different places also have high similarity values. Therefore, our model can learn more representative features with the proposed content aware pooling. Fig.~\ref{fig:teaser_b_upsample_vis} shows our predicted optical flow compared with other unsupervised methods. As can be seen, with the help of AFU, the interpolation can produce sharp motion boundaries. Equipped with CAP and AFU, the classical pyramid network has been upgraded, producing leading performance both quantitatively and qualitatively when evaluated on the flow benchmarks~\cite{Butler2012,KITTI_2012,KITTI_2015}. To sum up, our main contributions include:
	\vspace{-0.5em}
	
	\begin{itemize}
		\item We propose a Content Aware Pooling (CAP) module for the pyramid downampling. The CAP can assemble similar features locally, improving the capability of feature representation substantially. 
		\vspace{-0.5em}
		
		\item We propose an Adaptive Flow Upsampling (AFU) module for the pyramid upsampling, where the blurs caused by cross-edge interpolation can be avoided, yielding sharper motion boundaries.
		\vspace{-0.5em}
		
		\item We achieve superior performance over the state-of-the-art unsupervised methods, evaluated on multiple leading benchmarks.
		\vspace{-1em}
	\end{itemize}
	
	\begin{figure*}
		\centering
		\includegraphics[width=0.98\linewidth]{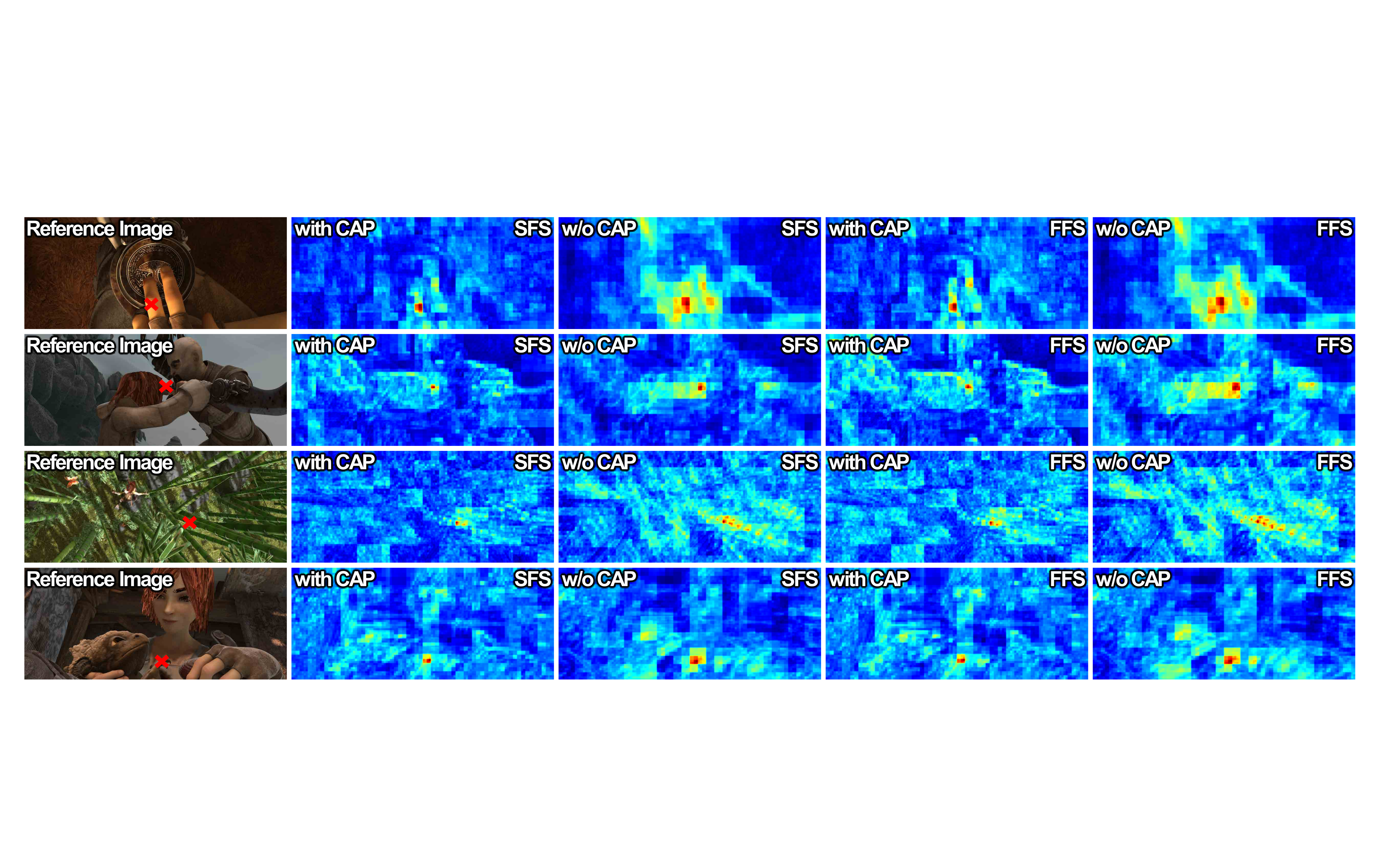}
		\caption{The feature matching visualizations of our CAP module vs. conventional striding in convolution. We extract features from the source and the target images. We pick a feature vector from the source feature map (red cross), and compute cosine differences with: other places in the source feature map (SFS), and with all features at the target feature map (FFS). More details are provided in Sec.~\ref{algo:adaptive_feature_grouping}. Red represents high similarity score and blue represents low similarity score. Features by SIC are likely to be similar with other places, while features by CAP is only similar with themselves.
		}\label{fig_algo_adaptive_feature_grouping_feature_vis}
		\vspace{-0.5em}
	\end{figure*}
	
	\section{Related Work}
	\subsection{Supervised Deep Optical Flow}
	Supervised methods require the annotated ground-truth flow labels to train the network~\cite{hui2020liteflownet3,zhao2020maskflownet}. FlowNet~\cite{Flownet_flyingchairs,FlowNet2} was first proposed by training on the flying chair dataset~\cite{Flownet_flyingchairs}. PWC-Net adopted the pyramid network that learns the motion from carose to fine, which calculates cost volumes at each pyramid level~\cite{pwc_net}. LiteFlowNet proposed to build lightweight networks for the efficiency~\cite{LiteFlowNet}. IRR-PWC propsoed an iterative residual refinement scheme in the pyramid network~\cite{irrpwc}. Recently, RAFT~\cite{raft2020} proposed to recurrently estimate flow fields on 4D correlation volumes, achieving state-of-the-art performance.
	
	\subsection{Unsupervised Deep Optical Flow}
	Unsupervised methods directly minimize the difference between two input images, by warping one to the other with predicted flow vectors. In this way, there is no need of ground-truth labels. However, the training becomes more difficult than supervised methods. Different methods with different focus have been proposed, including occlusion-aware losses by forward-backward check~\cite{unflow_2018aaai} and range-map occlusion check~\cite{wang2018}, census transform constrain~\cite{Ren2017aaai}, multi-frame formulation~\cite{unflow_multi_occ}, data argumentation~\cite{liu2020learning}, data distillation~\cite{Pengpeng2019, Liu2019CVPR}, epipolar constrain~\cite{Epipolar_flow_2019cvpr}, depth constrains~\cite{Anurag2019,dfnet_zou_2018} and feature similarity constrain~\cite{simFlow2020eccv}. By integrating multiple components, UFlow achieves the leading performance on multiple benchmarks~\cite{jonschkowski2020matters}. 
	
	\subsection{Image Guided Upsampling}
	Our method is also related to the edge-aware interpolation and upsampling, such as joint bilateral upsampling~\cite{kopf2007JBU} and guided image filtering~\cite{he2010guided}. Apart from the traditional methods, CNN approaches have also been attempted to extract guidance feature or guidance filter for upsampling~\cite{li2016fastfgi,wu2018fast,su2019pac}. We compare our AFU module with these opponents to demonstrate its effectiveness. 
	
	\begin{figure}
		\centering
		\includegraphics[width=0.9\linewidth]{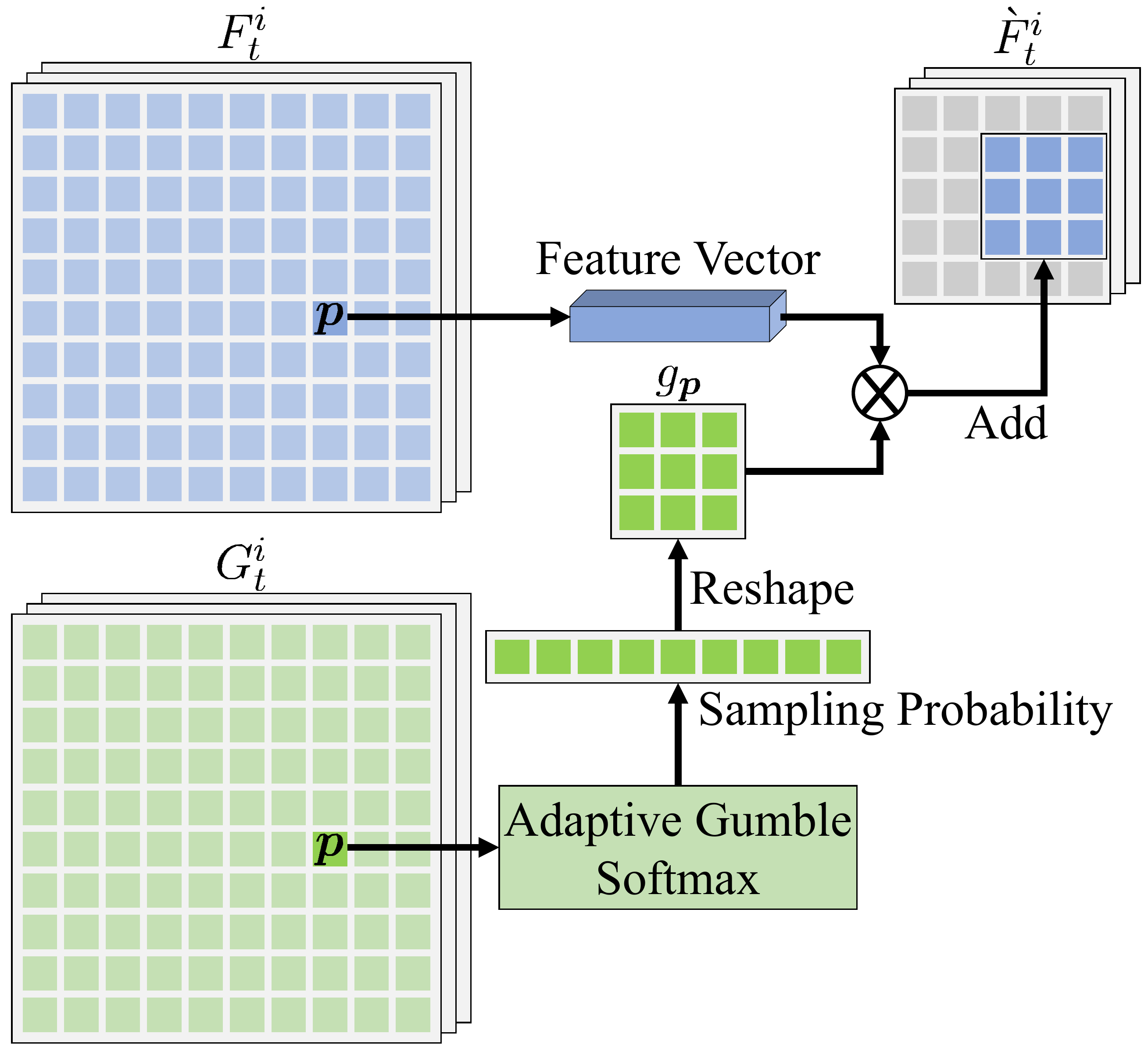}
		\caption{Illustration of our Content Aware Pooling module. For each feature vector in high resolution feature $F_{t}^{i}$, we add it to its corresponding neighbor position in low resolution feature $\grave{F}_{t}^{i}$ based on the sampling probability kernel $g_{\bm{p}}$ that is calculated by adaptive gumbel softmax and reshape operation. 
		}\label{fig_algo_adaptive_feature_grouping}
		\vspace{-0.5em}
	\end{figure}

	\begin{figure}
	\centering
	\includegraphics[width=0.85\linewidth]{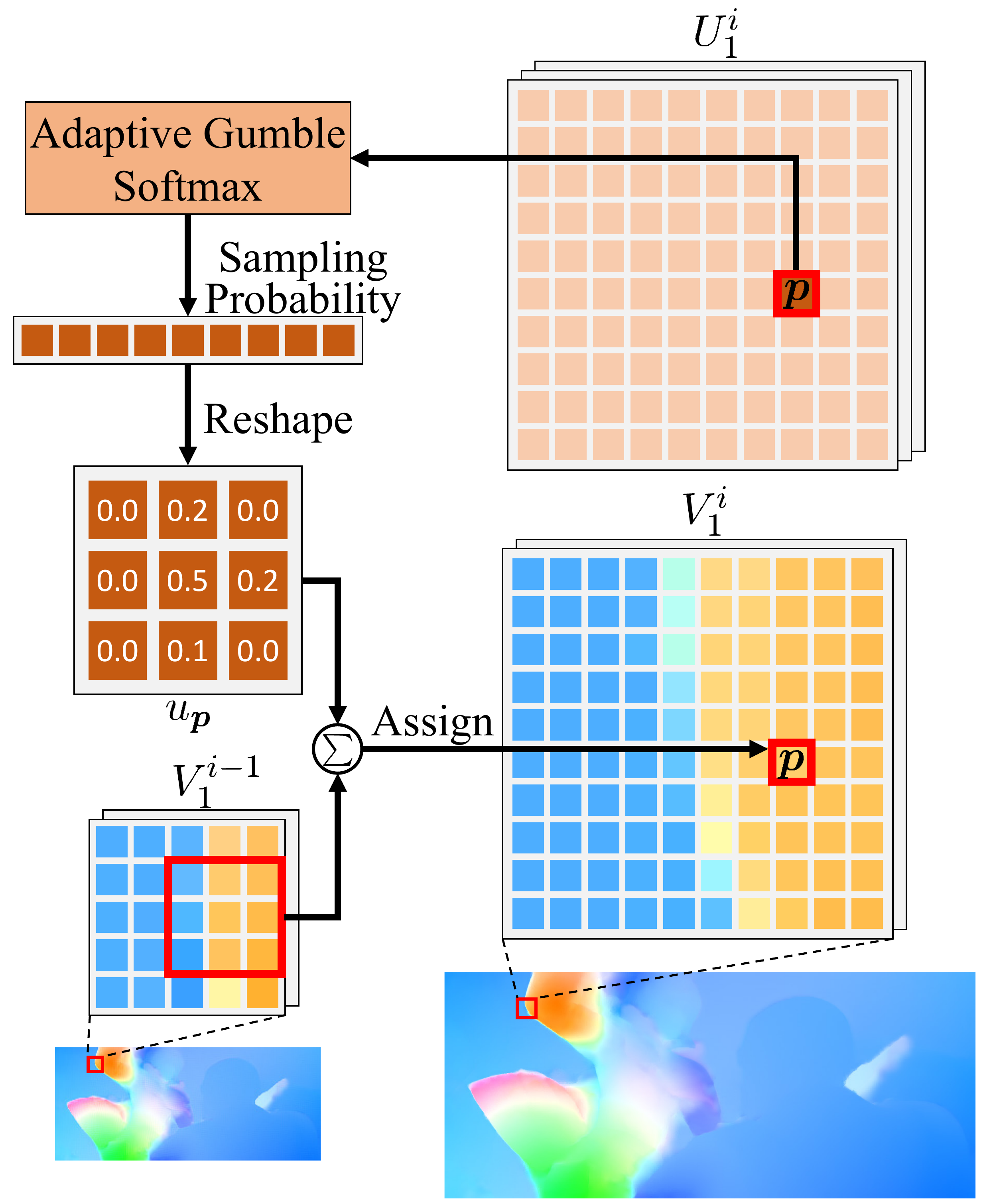}
	\caption{Illustration of our Adaptive Flow Upsampling module. The flow vector in high resolution flow field $V_1^{i}(\bm{p})$ is generated by sampling and fusion according to its sampling kernel $u_{\bm{p}}$.
	}\label{fig_algo_adaptive_flow_upsampling}
	\vspace{-0.5em}
	\end{figure}
	
	\section{Algorithm}	
	In this section, we first provide an overview of the network architecture of our method in Sec.~\ref{algo:overview}. Then we introduce the proposed Content Aware Pooling (CAP) module in Sec.~\ref{algo:adaptive_feature_grouping} and Adaptive Flow Upsampling (AFU) module in Sec.~\ref{algo:adaptive_flow_grouping}. Finally, we describe the loss functions used for unsupervised training in Sec.~\ref{algo:unsupervised_losses}.
	
	\subsection{Network Architecture}\label{algo:overview}
	The pipeline of the proposed network is illustrated in Fig.~\ref{fig_algo_our_pipeline}. It takes two frames $I_1$ and $I_{2}$ as inputs and produces an optical flow field $V_1$ that describes the motion of each pixel in $I_1$ towards $I_{2}$. The whole network contains three parts: an adaptive sub-net, a siamese feature encoder and a flow decoder. 
	
	First, we use the adaptive sub-net to extract multi-scale adaptive sampling maps which will be used later in the CAP module and the AFU module:
	\begin{align}
	\big\{G^{i}_{1}, G^{i}_{2}, U^{i}_{1}\big\} & = \mathcal{A}(I_1, I_2), \quad i \in \{0,1,..., N\}  \label{eq:adaptive_sub_net}
	\end{align}
	where $\mathcal{A}$ is our adaptive sub-net, $i$ is the index of each scale and small number represents the coarse scale, $G^{i}_{1}$, $G^{i}_{2}$ and $U^{i}_{1}$ are adaptive sampling maps. In our implementation, the adaptive sub-net is designed as a simple U-Net structure detailed in our supplementary files. 
	
	Second, in the siamese feature encoder, we extract multi-scale feature pairs from the input images to cover both global and local information, which is formulated as: 
	\begin{align}
	\grave{F}^{i}_{t} & = \mathcal{G}(F^{i}_{t},G^{i}_{t}), \label{eq:pyramid_feature_downscale}\\
	F^{i-1}_{t} & = \mathcal{C}^{i-1}(\grave{F}^{i}_{t}) \label{eq:pyramid_feature_conv}
	\end{align}
	where $t \in \{1, 2\}$ is index of the input images, $\mathcal{G}$ represents the proposed CAP module, $\grave{F}^{i}_{t}$ is the downsampled feature of $F^{i}_{t}$, and $\mathcal{C}^{i}$ is a convolution layer. 
	
	After the feature encoding process, we estimate flow fields by the flow decoder formulated as follows:
	\begin{align}
	\hat{V}_{1}^{i-1} & =\mathcal{U}(V_{1}^{i-1}, U_{1}^{i}), \label{eq:pyramid_flow_upsample}\\
	V_{1}^{i} & =\mathcal{D}(F_{1}^{i},F_{2}^{i},\hat{V}_{1}^{i-1}),  \label{eq:pyramid_flow_decoder}
	\end{align}
	where $\mathcal{U}$ represents our AFU module, $\hat{V}_{1}^{i-1}$ is the upsampled flow from $i-1$ scale and $\mathcal{E}$ is a flow estimator. Specifically, the flow estimator $\mathcal{D}$ is designed following the recent work UFlow~\cite{jonschkowski2020matters}, which contains feature warping, correlation layer, cost volume normalization, a dense convolution block and a dilated convolution block. 
	
	Generally, convolution layers with stride $=2$ are used to downscale feature maps. However, the regular downsampling method based on sliding windows may fuse features from different objects, reducing the matching accuracy of pair-wise correlation estimation. To tackle this issue, we propose CAP module to automatically group similar features in downsampling process, referred to as content aware pooling. Besides, we notice that the commonly used bilinear upsampling may introduce interpolation errors and blur artifacts during decoding process. Thus, the AFU module is proposed to ease this problem by adaptively interpolating flow fields with learnable weights. The details of these two modules are presented in Sec.~\ref{algo:adaptive_feature_grouping} and Sec.~\ref{algo:adaptive_flow_grouping}.
	
	
	\begin{figure*}[ht]
		\centering
		\includegraphics[width=1.0\linewidth]{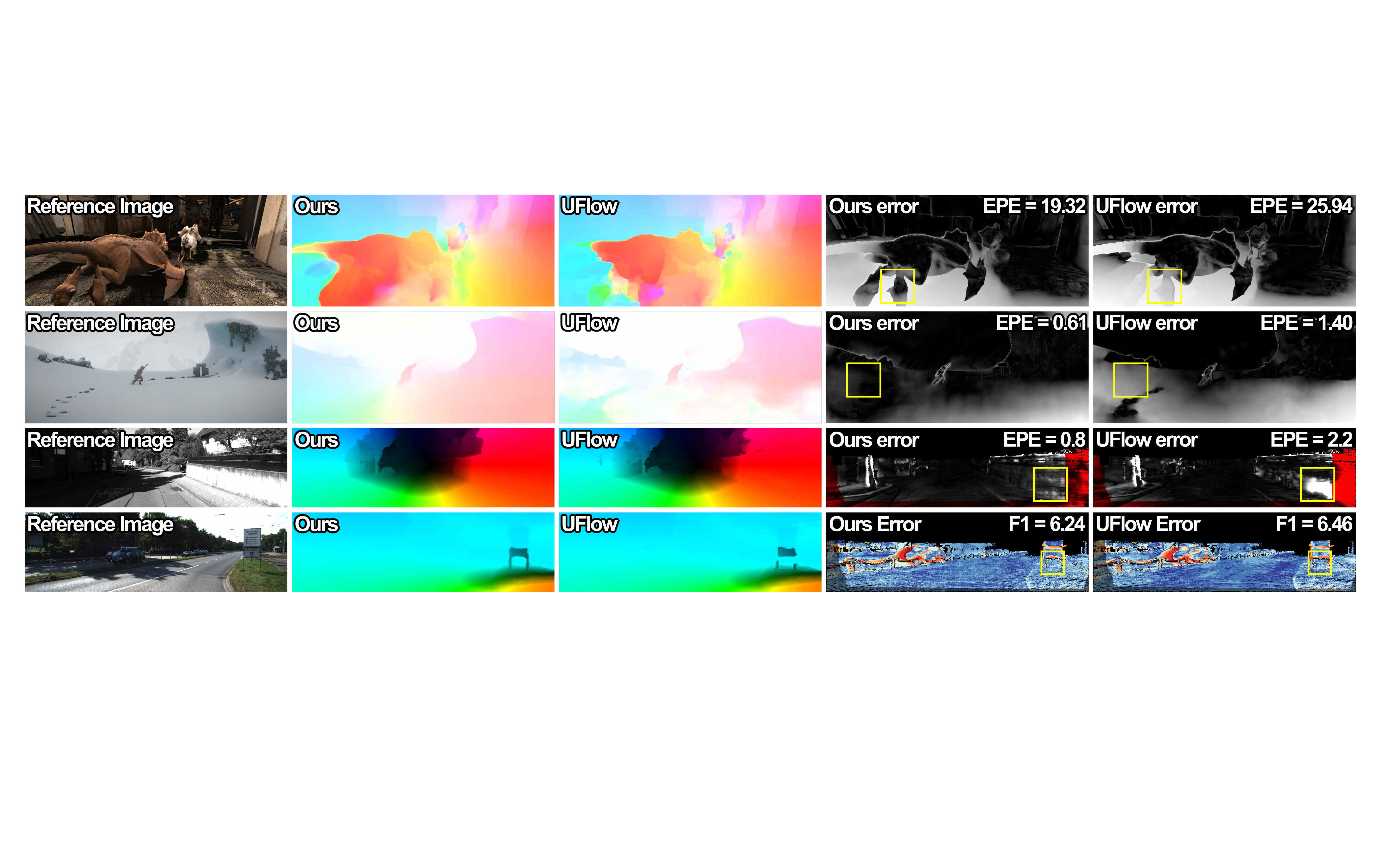}
		\caption{We show qualitative comparisons with the state-of-the-art method UFlow~\cite{jonschkowski2020matters} on online evaluation benchmarks, including Sintel Clean (first row), Final (second row), KITTI 2012 (third row) and 2015 (last row). The error maps of predictions are visualized in the last two columns. In error maps, brighter regions indicate the larger estimation errors except that visualized by KITTI 2015 benchmark where correct estimations are displayed in blue and wrong ones in red.}
		\label{fig:results_pk_benchmark}
		\vspace{-0.5em}
	\end{figure*}
	
	\subsection{Content Aware Pooling}\label{algo:adaptive_feature_grouping}
	As mentioned above, CAP module is proposed to automatically group similar features in the pooling process. The illustration of our CAP is shown in Fig.~\ref{fig_algo_adaptive_feature_grouping}. The input is a high resolution feature map $F_{t}^{i}$ with size of $H\times W\times c$, and an adaptive sampling map $G_{t}^{i}$ with size of $H\times W\times 11$, in which $9$ channels are used as sampling scores $\overline{G}_{t}^{i}$ and the rest 2 channels are used as the control parameter $\sigma$ and $\tau$ in our adaptive gumbel softmax. The output is a downsampled feature map $\grave{F}^{i}_{t}$ with size of $\frac{H}{r}\times \frac{W}{r}\times c$, where $c$ denotes the channel number and $r$ is the sampling rate, typically set to 2 in feature encoding process.
	
	For each feature vector $F_{t}^{i}(\bm{p})$ at spatial position $\bm{p}$, we calculate a sampling probability kernel $g_{\bm{p}}(\bm{q})$ from $G_{t}^{i}(\bm{p})$, which indicates the probability of $F_{i}^{t}(\bm{p})$ contributing to the neighbouring region of its corresponding position $\bm{q} \in \mathcal{N}(\bm{p}/r)$ in the low resolution feature $\grave{F}^{i}_{t}$. Then we generate the feature $\grave{F}^{i}_{t}$ by grouping and accumulating all feature vectors in $F_{t}^{i}$ according to their sampling probability (the `$\bigotimes$' and `Add' operation in Fig.~\ref{fig_algo_adaptive_feature_grouping}):
	\begin{align}
	\grave{F}^{i}_{t}(\bm{q}) & = \sum_{\bm{p}\in \mathcal{N}_{\times r}(\bm{q})}{g_{\bm{p}}(\bm{q})F_{t}^{i}(\bm{p})}, \label{eq:adaptive_feature_grouping_method}
	\end{align}
	where $\mathcal{N}_{\times r}(\bm{q})$ is a set of pixels in $F_{t}^{i}$ whose sampling probability kernel covers position $\bm{q}$ in  $\grave{F}^{i}_{t}$. 
	
	In order to avoid feature grouping across different regions, we use adaptive gumbel softmax~\cite{ojha2019elastic,wang2019prnet} to suppress small probabilities when producing sampling probability kernels. The adaptive sampling map $G_{t}^{i}$ is first splitted as sampling scores $\overline{G}_{t}^{i}(j,\bm{p})$ and control parameters $\sigma(\bm{p})$ and $\tau(\bm{p})$ to control the distribution tendency of sampling kernels, where $j$ is channel index and $\bm{p}$ is spatial coordinate. In summary, the adaptive gumbel softmax can be formulated as follows:
	\begin{align}
	x(j, \bm{p}) & = \frac{\overline{G}_{t}^{i}(j,\bm{p})-|\sigma(\bm{p})|}{\text{sigmoid}(\tau(\bm{p}))+\rho},  \label{eq:adaptive_gumbel_softmax_control}\\
	k_{\bm{p}}(j) & =\frac{exp(x(j, \bm{p}))}{\sum_{k}^{9} exp(x(k, \bm{p}))},  \label{eq:adaptive_gumbel_softmax}
	\end{align}
	where $\rho$ is a constant to avoid zero denominator and $x(j, \bm{p})$ is the transformed sampling score.
	
	Fig.~\ref{fig_algo_adaptive_feature_grouping_feature_vis} provides some visualizations of content aware pooling results by comparing our CAP module with conventional striding in convolution (SIC). We first interpolate pyramid features into the image size and concatenate them together. Then feature vector in $I_1$ located by the red cross is selected to calculate cosine similarity with features of $I_1$ and $I_2$, which is the self feature similarity (SFS) map and the forward feature similarity (FFS) map, respectively. The SFS map reveals the discriminative ability of the encoded features and the FFS map reveals the matching ability between feature pairs. From Fig.~\ref{fig_algo_adaptive_feature_grouping_feature_vis}, we can see that feature extracted by SIC method is likely to be similar with neighbor objects, while feature by our CAP module is only similar with its corresponding feature vector. 
	
	\begin{table*}[t]
		\centering
		\resizebox*{0.90 \textwidth}{!}{
			\begin{tabular}{
					>{\centering\arraybackslash}p{0.3cm}
					p{3.2cm} 
					||>{\centering\arraybackslash}p{0.8cm} 
					|>{\centering\arraybackslash}p{0.8cm} 
					|>{\centering\arraybackslash}p{0.8cm} 
					|>{\centering\arraybackslash}p{2.2cm} 
					|>{\centering\arraybackslash}p{0.8cm} 
					|>{\centering\arraybackslash}p{0.8cm} 
					|>{\centering\arraybackslash}p{0.8cm} 
					|>{\centering\arraybackslash}p{0.8cm} 
				}
				\toprule
				\multicolumn{2}{c}{\multirow{2}{*}{Method}} & \multicolumn{2}{c}{KITTI 2012} & \multicolumn{2}{c}{KITTI 2015} & \multicolumn{2}{c}{Sintel Clean}&\multicolumn{2}{c}{Sintel Final}\\
				\cmidrule(lr){3-4} \cmidrule(lr){5-6} \cmidrule(lr){7-8} \cmidrule(lr){9-10}
				&&train & test &train & test (F1-all) &train &test &train &test
				\\
				\midrule
				\multirow{9}{*}{\rotatebox{90}{Supervised}}
				& FlowNetS~\cite{Flownet_flyingchairs}    &8.26  &  --  &  --  &   --  & 4.50 &7.42  &5.45  &8.43  \\
				& FlowNetS+ft~\cite{Flownet_flyingchairs} &7.52  &9.1   &  --  &   --  &(3.66)&6.96  &(4.44)&7.76  \\
				& SpyNet~\cite{spynet2017}                &9.12  &  --  &  --  &   --  &4.12  &6.69  &5.57  &8.43  \\
				& SpyNet+ft~\cite{spynet2017}             &8.25  &10.1  &  --  &35.07\%&(3.17)&6.64  &(4.32)&8.36  \\
				& LiteFlowNet~\cite{LiteFlowNet}          &4.25  &  --  &10.46 &   --  & 2.52 & -- & 4.05 &--  \\
				& LiteFlowNet+ft~\cite{LiteFlowNet}       &(1.26)& 1.7  &(2.16) &10.24\%&(1.64)& 4.86 &(2.23)&6.09  \\
				& PWC-Net~\cite{pwc_net}                  & 4.57 &  --  &13.20 &   --  & 3.33 &  --  & 4.59 &  --  \\
				& PWC-Net+ft~\cite{pwc_net}               &(1.45)& 1.7  &(2.16)&9.60\% &(1.70)& 3.86 &(2.21)&5.13  \\
				& IRR-PWC+ft~\cite{irrpwc}                &  --  &  --  &(1.63)&7.65\% &(1.92)& 3.84 &(2.51)&4.58  \\
				& RAFT~\cite{raft2020}                    &  --  &  --  &5.54& -- &1.63& -- &2.83 & --  \\
				& RAFT-ft~\cite{raft2020}                 &  --  &  --  &--& 6.30\% & -- & 2.42 & -- & 3.39  \\
				\midrule
				\midrule
				\multirow{12}{*}{\rotatebox{90}{Unsupervised}}
				& BackToBasic~\cite{Jason2016}               & 11.30& 9.9 &  --  &   --  &  --  &  --  &  --  &  --  \\
				& DSTFlow~\cite{Ren2017aaai}                 & 10.43& 12.4& 16.79& 39\%  &(6.16)&10.41 &(6.81)&11.27 \\
				& UnFlow~\cite{unflow_2018aaai}              & 3.29 &  --  & 8.10 & 23.3\%&  --  &9.38  &(7.91)&10.22 \\
				& OAFlow~\cite{wang2018}                     & 3.55 & 4.2 & 8.88 & 31.2\%&(4.03)&7.95  &(5.95)&9.15  \\
				& Back2Future~\cite{unflow_multi_occ}        &  --  &  --  & 6.59 &22.94\%&(3.89)&7.23  &(5.52)&8.81  \\
				& NLFlow~\cite{tip2020_nonlocalflow}         &3.02  &4.5   &6.05  &22.75\%&(2.58)&7.12  &(3.85)&8.51  \\
				& DDFlow~\cite{Pengpeng2019}                 &2.35  &3.0   &5.72  &14.29\%&(2.92)&6.18 &(3.98)&7.40  \\
				& EpiFlow~\cite{Epipolar_flow_2019cvpr}      &(2.51)&3.4   &(5.55)&16.95\%&(3.54)&7.00  &(4.99)&8.51  \\
				& SelFlow~\cite{Liu2019CVPR}                 &1.69  &2.2   &4.84  &14.19\%&(2.88)&6.56  &(3.87)&6.57  \\
				& STFlow~\cite{tip2020_nonlocalflow}         &1.64  &1.9   &3.56  &13.83\%&(2.91)&6.12  &(3.59)&6.63  \\
				& ARFlow~\cite{liu2020learning}              &\textcolor{blue}{1.44} &\textcolor{blue}{1.8} &2.85  &11.80\%&(2.79)&\textcolor{blue}{4.78}   &(3.87)&\textcolor{blue}{5.89}   \\
				& SimFlow~\cite{simFlow2020eccv}        & --  & --   &5.19  &13.38\%&(2.86)&5.92  &(3.57)&6.92  \\
				& UFlow~\cite{jonschkowski2020matters}       &1.68  &  1.9 &\textcolor{blue}{2.71}  &\textcolor{blue}{11.13\%}&\textcolor{blue}{ (2.50)}&5.21  &\textcolor{blue}{(3.39)} &6.50  \\
				\midrule
				& ASFlow(ours)                                       & \bf{1.26} & \bf{1.5} & \bf{2.47} & \bf{9.67\%}&\bf{(2.40)}&\bf{4.56}  &\bf{(2.89)}&\bf{5.86}  \\
				\bottomrule
			\end{tabular}
		}
		\vspace{0.5em}
		\caption{Quantitative comparison with state-of-the-art methods on four widely-used datasets using EPE and F1-measure metrics (the lower the better). Following previous works~\cite{jonschkowski2020matters, simFlow2020eccv, liu2020learning}, `$-$' means the result is not reported in the paper, `$(\ )$' indicates images from test set are used during unsupervised training, and `+ft' means the supervised methods use images of target domain for training, otherwise using synthetic data like Flying Chairs~\cite{Flownet_flyingchairs} and Flying Chairs occ~\cite{irrpwc}. The best unsupervised method is marked in {\bf bold} and the second best is marked in {\color{blue} blue} for better comparison. 
		}
		\label{tab:SOTAs}
	\end{table*}
	
	\subsection{Adaptive Flow Upsampling}\label{algo:adaptive_flow_grouping}
	The conventional bilinear upsampling method may interpolate flow vectors across object boundaries leading to blur artifacts and errors during flow decoding process. To solve this problem, we design an adaptive flow upsampling module to adaptively interpolate flow fields with learnable weights. The detail of our AFU module is shown in Fig.~\ref{fig_algo_adaptive_flow_upsampling}. Given a low resolution flow field $V_{1}^{i-1}$ of size $\frac{H}{r}\times \frac{W}{r} \times 2$ and a high resolution adaptive sampling map $U_{1}^{i}$ with size of $H\times W \times 11$, our goal is to produce a high resolution flow field $V_{1}^{i}$ with size of $H\times W \times 2$. We define $\bm{p}$ as a spatial coordinate in $V_{1}^{i}$ and $\bm{q} \in \mathcal{N}(\bm{p}/r)$ as its corresponding neighbors in $V_{1}^{i-1}$. 
	The flow vectors in high resolution flow field $V_{1}^{i}$ can be calculated by the following formulation (the `$\sum$' and `Assign' operation in Fig.~\ref{fig_algo_adaptive_flow_upsampling}):
	\begin{align}
	V_{1}^{i}(\bm{p}) & = \sum_{\bm{q}\in \mathcal{N}(\bm{p}/r)}{u_{\bm{p}}(\bm{q})V_{1}^{i-1}(\bm{q})},  \label{eq:adaptive_flow_upsampling}
	\end{align}
	where $u_{\bm{p}}(\bm{q})$ is a sampling probability kernel generated from $U_{1}^{i}$ to indicate the contribution probability of $V_{1}^{i-1}(\bm{q})$ to $V_{1}^{i}(\bm{p})$. 
	The flow vectors in high resolution flow field is generated by adaptively fusing flow vectors in low resolution flow field based on sampling probability kernels. Note that, in order to suppress the probability of flow fusion across edges, we use adaptive gumbel softmax as in Eq.~\ref{eq:adaptive_gumbel_softmax_control} and Eq.~\ref{eq:adaptive_gumbel_softmax} to produce the kernels, where small probabilities are compressed to zeros. 
	
	\subsection{Unsupervised Losses}\label{algo:unsupervised_losses}
	In order to train our network in unsupervised setting where ground-truth is not available, we use a set of unsupervised losses as our training objective. 
	Our main objective is the photometric loss $\mathcal{L}_{d}$, which is designed based on the brightness constancy assumption that the object appearance should be invariable in input frames. However, occlusion regions caused by moving objects can not be optimized by the photometric loss. We explicitly exclude these regions in the photometric loss by forward-backward consistency checking~\cite{unflow_2018aaai}. As a result, the photometric loss $\mathcal{L}_{d}$ is formulated as follows:
	\begin{align}
	\mathcal{L}_{d}=\frac{\sum_{\bm{p}}{\Psi\Big(I_1(\bm{p})-I_{2}\big(\bm{p}+V_{1}(\bm{p})\big)\Big) \cdot O_{1}(\bm{p})}}{\sum_{\bm{p}}{O_{1}(\bm{p})}}, \label{eq:data_term_loss} 
	\end{align}
	where $O_{1}$ is the occlusion mask generated by forward-backward consistency checking. `$1$' indicates the non-occluded pixel and `$0$' means the occluded pixel. $\Psi$ is the robust penalty function~\cite{Pengpeng2019}: $\Psi(x)=(|x|+\epsilon)^q$ in which $q$ and $\epsilon$ are set to 0.4 and 0.01. 
	
	Following previous works, several loss functions are used to train our model, including the edge-aware smooth loss $\mathcal{L}_s$ that improves the smoothness of output flow field~\cite{wang2018}, the census loss $\mathcal{L}_c$ that increases the robustness under illumination changes~\cite{unflow_2018aaai}, the boundary dilated warping loss $\mathcal{L}_{b}$ to learn motions towards outside the image plane~\cite{luo2020occinpflow}, the augmentation regularization loss $\mathcal{L}_a$ that introduces the equivariance constrain to encourage the robustness to variations~\cite{liu2020learning}. 
	
	In order to ensure the upsampled flow fields to better fit object boundaries, we design a sampling regularization loss $\mathcal{L}_{r}$ to constrain the learned adaptive sampling maps $\{U_1^{i}\}$. We first downscale the input image $I_1$ to $I_1^0$, whose size is the same as $V_1^0$. Then we iteratively upsample the downscaled image and compute a reconstruction loss with the original image, which is formulated as follows:
	\begin{align}
	I_1^{i} & =\mathcal{U}(I_1^{i-1},U_1^{i}), \label{eq:sampling_regularization_image_reconstruction} \\
	\mathcal{L}_{r} & =\sum_{\bm{p}}{\Psi(I_1(\bm{p})-I_1^{N}(\bm{p}))}, \label{eq:sampling_regularization_loss} 
	\end{align}
	where $I_1^{N}$ is the reconstructed image by the iterative upsampling process described in Eq.~\ref{eq:sampling_regularization_image_reconstruction}. 
	
	Eventually, our loss function is a weighted combination of above individual loss terms:
	\begin{eqnarray}\label{eq:total-training-loss}
	\mathcal{L}=\mathcal{L}_d+\lambda_s\mathcal{L}_s+\lambda_c\mathcal{L}_c+\lambda_a\mathcal{L}_a+\lambda_r\mathcal{L}_r, 
	\end{eqnarray}
	where $\lambda_s$, $\lambda_c$, $\lambda_a$ and $\lambda_r$ are hyper-parameters, set to $\lambda_s=0.05$, $\lambda_c=1$, $\lambda_a=0.5$, $\lambda_r=0.1$ in our experiments. 
	
	\begin{figure*}
		\centering
		\includegraphics[width=0.98\textwidth]{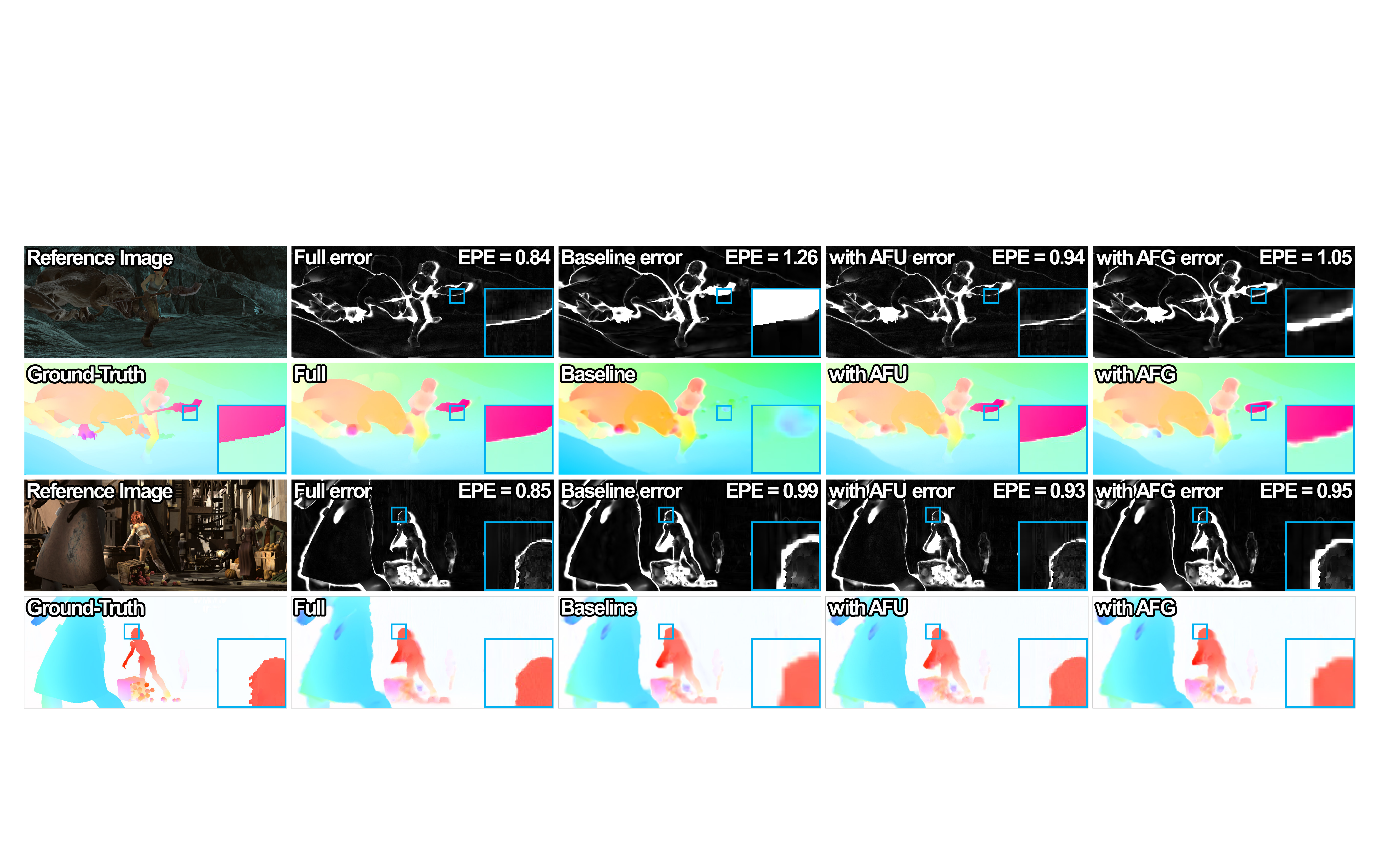}
		\caption{Qualitative visualizations of the proposed method on Sintel Clean. The room in flows and error maps are shown in the right corner of each sample.  
		}\label{fig:ablation}
	\vspace{-0.5em}
	\end{figure*}
	\begin{table*}[pt]
	\centering
	\resizebox*{0.98 \textwidth}{!}{
		\begin{tabular}{
				>{\centering\arraybackslash}p{0.8cm} 
				>{\centering\arraybackslash}p{0.8cm} 
				>{\centering\arraybackslash}p{0.8cm} 
				>{\centering\arraybackslash}p{0.8cm} 
				>{\centering\arraybackslash}p{0.8cm} 
				||>{\centering\arraybackslash}p{0.8cm} 
				|>{\centering\arraybackslash}p{0.8cm} 
				|>{\centering\arraybackslash}p{0.8cm} 
				|>{\centering\arraybackslash}p{0.8cm} 
				|>{\centering\arraybackslash}p{0.8cm} 
				|>{\centering\arraybackslash}p{0.8cm} 
				|>{\centering\arraybackslash}p{0.8cm} 
				|>{\centering\arraybackslash}p{0.8cm} 
				|>{\centering\arraybackslash}p{0.8cm} 
				|>{\centering\arraybackslash}p{0.8cm} 
				|>{\centering\arraybackslash}p{0.8cm} 
				|>{\centering\arraybackslash}p{0.8cm} 
			}
			\toprule
			\multirow{2}{*}{CL} & \multirow{2}{*}{BDWL} & \multirow{2}{*}{ARL} & \multirow{2}{*}{CAP} & \multirow{2}{*}{AFU} & \multicolumn{3}{c}{KITTI 2012} & \multicolumn{3}{c}{KITTI 2015} & \multicolumn{3}{c}{Sintel Clean} & \multicolumn{3}{c}{Sintel Final} \\
			\cmidrule(lr){6-8} \cmidrule(lr){9-11} \cmidrule(lr){12-14} \cmidrule(lr){15-17}
			&                      &                      &                      &                      & ALL      & NOC         & OCC         & ALL         &    NOC      &    OCC      &    ALL       &    NOC       &    OCC      &    ALL       &    NOC       &    OCC      \\
			\midrule
			&                      &                      &                      &                      &   4.52   &    1.76     &   19.63     &   7.58      &   2.46      &  30.43      &   (3.52)       &   (1.87)       &  (12.9)      &   (4.19)       &   (2.59)       &  (13.64)      \\
			\Checkmark      &                      &                      &                      &                      &   3.39   &    1.09     &   16.58     &   6.89      &   2.20      &  28.12      &   (3.41)       &   (1.62)       &  (13.5)      &   (3.85)       &   (2.17)       &  (13.71)      \\
			\Checkmark      &     \Checkmark       &                      &                      &                      &   1.42   &    0.91     &   4.39      &   3.00      &   2.12      &   6.89      &   (2.84)       &   (1.50)       &  (10.6)      &   (3.60)       &   (2.28)       &  (11.52)      \\
			\Checkmark      &     \Checkmark       &     \Checkmark       &                      &                      &   1.37   &    0.93     &   3.98      &   2.64      &   1.96      &   6.01      &   (2.61)       &   (1.33)       &  (10.1)      &   (3.17)       &   (1.92)       &  (10.70)      \\
			\Checkmark      &     \Checkmark       &     \Checkmark       &     \Checkmark       &                      &   1.29   &    0.89     &   3.78      &   2.53      &   1.98      &   5.16      &   (2.51)       &   (1.27)       &   (9.79)      &   (2.98)       &   (1.79)       &  (9.98)      \\
			\Checkmark      &     \Checkmark       &     \Checkmark       &                      &     \Checkmark       &   1.30   &    0.88     &   3.82      &   2.57      &   1.99      &   5.08      &   (2.46)       &   (1.23)       &  (9.63)      &   (2.94)       &   (1.73)       &  (10.07)      \\
			\Checkmark      &     \Checkmark       &     \Checkmark       &     \Checkmark       &     \Checkmark    &\textbf{1.26}&\textbf{0.87}&\textbf{3.72}&\textbf{2.47}&\textbf{1.93}&\textbf{5.02}&\textbf{(2.40)} &\textbf{(1.20)} &\textbf{(9.36)}&\textbf{(2.89)} &\textbf{(1.71)} &\textbf{(9.89)}\\
			\bottomrule
		\end{tabular}
	}
	\vspace{0.5em}
	\caption{Ablation for unsupervised components. CL: census loss~\cite{unflow_2018aaai}, BDWL: boundary dilated warping loss~\cite{luo2020occinpflow}, ARL: augmentation regularization loss~\cite{liu2020learning}, SGU: self-guided upsampling, PDL: pyramid distillation loss. The best results are marked in {\bf bold}.
	}
	\label{tab:ablation_all}
	\end{table*}
	
	\section{Experimental Results}
	
	\subsection{Datasets and Implementation Details} 
	We conduct comprehensive experiments on four widely-uesd optical flow benchmarks, including MPI-Sintel~\cite{Butler2012}, KITTI 2012~\cite{KITTI_2012}, and KITTI 2015~\cite{KITTI_2015}. MPI-Sintel contains 1,041 training image pairs extracted from the rendered open-source movie, divided into `Clean' and `Final' passes. Following previous works~\cite{jonschkowski2020matters, simFlow2020eccv, liu2020learning}, we use both versions of rendering images to train our model. For KITTI 2012 and 2015, we first use the 28,058 image pairs from KITTI raw dataset to pre-train the model, and then perform finetuning on multi-view extension data.
	
	The implementation of the proposed ASFlow is based on PyTorch toolbox. We train our model on 2 NVIDIA GeForce GTX 2080Ti GPUs for about 1000k iterations. For better generalization, we follow previous work~\cite{liu2020learning} to use basic data augmentation strategies like random crop and horizontal flip for training. The standard evaluation metrics, i.e., average endpoint error (EPE) and the percentage of erroneous pixels (F1-measure), are used to evaluate the performance of the predicted optical flow.
	
	\subsection{Comparison with State-of-the-Arts} 
	
	In Tab.~\ref{tab:SOTAs}, We compare our method with State-of-the-Art (SOTA) works, including both of supervised and unsupervised methods, on four widely-used datasets. The best unsupervised method is marked in {\bf bold} and the second best is marked in {\color{blue} blue} for better comparison.
	
	\vspace{-1.2em}
	\paragraph{Comparison with Unsupervised Methods.}  
	As shown in Tab.~\ref{tab:SOTAs}, our ASFlow consistently achieves better performance than other methods on four standard benchmarks. Specifically, our method achieves an EPE error of 1.5 on KITTI 2012 test set, which surpasses previous top-ranked methods UFlow~\cite{jonschkowski2020matters} and ARFLow~\cite{liu2020learning} by around 21.1\% (1.9 $\rightarrow$ 1.5) and 16.7\% (1.8 $\rightarrow$ 1.5), respectively. For KITTI 2015 online evaluation, our method set new records of 2.47 in EPE on training set and 9.67\% in F1-measure, which outperforms previous methods by a large margin. On the most challenging dataset MPI-Sintel, our method achieves EPE scores of 4.56 on `Clean' pass for online testing. It obtains EPE = 5.86 on `Final' pass, outperformimg previous top methods SimFlow~\cite{simFlow2020eccv} and UFlow~\cite{jonschkowski2020matters} by 1.06 and 0.64 in terms of EPE. It is worth noting that our method is the first one to achieve the best results on all benchmarks, as shown in each line of Tab.~\ref{tab:SOTAs} (best viewed in colors).
	
	Fig.~\ref{fig:results_pk_benchmark} provides some qualitative comparisons with the previous best method UFlow~\cite{jonschkowski2020matters}. As can be seen, our method is clearly able to make accurate and smooth predictions, especially when handling the tough regions around foreground boundary.
	
	\vspace{-1.2em}
	\paragraph{Comparison with Supervised Methods.}
	We also report the results of representative supervised methods for comprehensive comparison, see Tab.~\ref{tab:SOTAs}. For cross domain evaluation, we consider the ground-truth of optical flow is not available for training. Thus, the supervised models are trained on synthetic data such as Flying Chairs~\cite{Flownet_flyingchairs} and Flying Chairs occ~\cite{irrpwc}, while the training procedure of the unsupervised methods can be directly performed only using target domain images. As can be seen, our method achieves better performance than all the supervised methods. Especially in real scenarios like KITTI 2015 dataset, it significantly outperforms the well-known supervised methods like LiteFlowNet~\cite{LiteFlowNet}, PWC-Net~\cite{pwc_net} and RAFT~\cite{raft2020} by a large margin (7.99, 10.73 and 3.07 in EPE, respectively).
	
	As for in-domain evaluation, our method generally achieve competitive performance with the supervised methods. Specially, on KITTI 2012 and 2015 datasets, our method achieves 1.5 in EPE and 9.67\% in F1-measure, which surprisingly exceed the recent supervised method like and LiteFlowNet~\cite{LiteFlowNet}.
	
	\subsection{Ablation Study} 
	In this section, we conduct a series of ablation expriments to evaluate each component in the proposed network. Following~\cite{Liu2019CVPR,simFlow2020eccv}, we train our model on train sets of KITTI and MPI-Sintel. The EPE error over all pixels (ALL), non-occluded pixels (NOC) and occluded pixels (OCC) are reported for quantitative comparisons.
	
	\vspace{-1.2em}
	\paragraph{Unsupervised Components.}
	Following the success of prior works~\cite{luo2020occinpflow, unflow_2018aaai, luo2020occinpflow}, we employ some effective components to boost the training of our model in an unsupervised manner. As shown in the first line of Tab.~\ref{tab:ablation_all}, we first train a baseline model using photometric loss and smooth loss, without the proposed modules. After adding census loss~\cite{unflow_2018aaai} (CL), boundary dilated warping loss~\cite{luo2020occinpflow} (BDWL) and augmentation regularization loss (ARL), it obtains consistent improvements by three metrics on all datasets, which demonstrates these three modules benefit to boosting a better prediction. Meanwhile, the performance of this model (CL + BDWL + ARL) is equivalent to that reported in previous best method UFlow~\cite{jonschkowski2020matters}. In addition, replacing the original striding strategy by our CAP in the each stage of encoder network greatly improves the performance. Similarly, we append our AFU module on decoders, and observe that the three metrics are clearly reduced (the lower the better). Finally, we fully equip the model with both of CAP and AFU, which brings about 10\% performance improvement.
	
	\begin{table}[!t]
		\centering
		\resizebox*{1.0\linewidth}{!}{
			\begin{tabular}{
					p{1.4cm} 
					||>{\centering\arraybackslash}p{1.8cm} 
					|>{\centering\arraybackslash}p{1.8cm} 
					|>{\centering\arraybackslash}p{1.8cm} 
					|>{\centering\arraybackslash}p{1.8cm} 
				}
				\toprule
				Method              &  KITTI 2012  &   KITTI 2015 & Sintel Clean  & Sintel Final  \\
				\midrule
				Bilinear                             &     1.29     &     2.53     &   (2.51)        &   (2.98)       \\
				JBU~\cite{kopf2007JBU}               &     1.51     &     3.00     &   (2.66)        &   (2.98)      \\
				GF~\cite{he2010guided}               &     1.40     &     2.90     &   (2.72)        &   (2.92)       \\
				\cmidrule(lr){1-5}
				DJF~\cite{li2016deep}                &     1.36     &     2.79     &   (2.75)        &   (3.20)       \\
				DGF~\cite{wu2018fast}                &     1.41     &     3.14     &   (2.69)        &   (3.05)       \\
				PAC~\cite{su2019pac}                 &     1.42     &     2.65     &    (2.58)       &   (2.95)      \\
				AFU                               &     1.28     &     2.52     &    (2.45)       &   (2.90)       \\
				AFU-RL                              & \textbf{1.26}& \textbf{2.47}& \textbf{(2.40)} & \textbf{(2.89)} \\
				\bottomrule
		\end{tabular}}
		\caption{Comparison of our AFU with classical upsampling methods, such as JBU~\cite{kopf2007JBU} and GF~\cite{he2010guided}, and deep-based upsampling methods, such as DJF~\cite{li2016deep}, DGF~\cite{wu2018fast} and PAC~\cite{su2019pac}. AFU-RL denotes the sampling regularization loss is used to enable the upsampled flow to better fit object boundaries.
		}
		\label{tab:upsampling}
		\vspace{-0.5em}
	\end{table}
	
	\begin{table}[!t]
		\centering
		\resizebox*{1.0\linewidth}{!}{
			\begin{tabular}{
					p{1.4cm} 
					||>{\centering\arraybackslash}p{1.8cm} 
					|>{\centering\arraybackslash}p{1.8cm} 
					|>{\centering\arraybackslash}p{1.8cm} 
					|>{\centering\arraybackslash}p{1.8cm} 
				}
				\toprule
				Method              &  KITTI 2012  &   KITTI 2015 & Sintel Clean  & Sintel Final  \\
				\midrule
				Bilinear          &     1.51     &     2.81     &   (2.75)        &   (3.20)       \\
				AVE               &     1.39     &     2.75     &   (2.66)        &   (2.98)      \\
				MAX               &     1.40     &     2.69     &   (2.72)        &   (3.02)       \\
				SIC               &     1.30     &     2.57     &   (2.46)        &   (2.94)       \\
				CAP            & \textbf{1.26}& \textbf{2.47}& \textbf{(2.40)} & \textbf{(2.89)} \\
				\bottomrule
		\end{tabular}}
		\caption{Comparison of our CAP with different feature pooling methods: average pooling (AVE), max pooling (MAX), and striding in convolution (SIC).
		}
		\label{tab:pooling}
		\vspace{-0.5em}
	\end{table}
	
	\vspace{-1.2em}
	\paragraph{Ablation for Upsampling Modules.}
	There have been several works attempt to propose general upsampling operations based on image information, such as JBU~\cite{kopf2007JBU}, GF~\cite{he2010guided}, DJF~\cite{li2016deep}, DGF~\cite{wu2018fast} and PAC~\cite{su2019pac}. However, these methods are not suitable to this challenging task. Here we propose a task specific upsampling strategy to better serve the need of optical flow upsampling. To verify the effect of our method, we carry out extensive comparisons with the upasampling methods. Specifically, we build a simple pyramid network with the same loss function, and repetitively change upsampling operations with the modules mentioned above for fair comparison. As we can see in Tab.~\ref{tab:upsampling}, our AFU achieves the best performance over all the competitors. This is because AFU can adaptively interpolate flow fields with learnable weights in pyramid decoders, so that the blur artifacts caused by cross-edge interpolation can be avoided, see column 4 of Fig.~\ref{fig:ablation}.
	
	\vspace{-1.2em}
	\paragraph{Ablation for Feature Pooling Strategies.}
	Tab.~\ref{tab:pooling} reports the comparison of our CAP with typical pooling strategies, including average pooling (AVE), max pooling (MAX), and striding in convolution (SIC). For fair comparison, all the experiments are conducted under the same setting. As we can see, our CAP consistently obtain better scores than others on four datasets. As mentioned in Sec.~\ref{algo:adaptive_feature_grouping}, the features are adaptively grouped based on content and appearance similarity, which helps the network to maintain spatial details of different objects. Experimental results demonstrate the obtained distinctive information is crucial for recovering the optical flow on thin stuffs as shown in Fig.~\ref{fig:ablation} (first sample, column 3 and 5).
	
	\vspace{-0.5em}
	\section{Conclusion}
	\vspace{-0.5em}
	We have presented ASFlow, an adaptive pyramid sampling method for unsupervised optical flow estimation. Two modules have been proposed, content aware pooling (CAP) for the pyramid downsampling and adaptive flow upsampling (AFU) for the upsampling. We compare our method with previous representative optical flow methods on the several leading benchmarks. In the further, we will explore the proposed two modules in the other applications, especially the CAP for the high-level vision tasks.


\end{document}